\documentclass{clv3}

\usepackage{hyperref}
\usepackage{xcolor}
\usepackage{times}
\usepackage{latexsym}

\usepackage{url}

\usepackage{algorithm} 
\usepackage{algorithmic} 
\usepackage{multirow} 
\usepackage{amsmath}
\usepackage{xcolor}
\usepackage{epsfig}
\usepackage{array}
\usepackage{bbding}
\usepackage{threeparttable}
\usepackage{CJK}

\definecolor{darkblue}{rgb}{0, 0, 0.5}
\hypersetup{colorlinks=true,citecolor=darkblue, linkcolor=darkblue, urlcolor=darkblue}

\issue{1}{1}{2017}

\runningtitle{A Semantic Relevance Based Neural Network for Summarization and Simplification}

\runningauthor{Shuming Ma et al.}

\begin{document}
\begin{CJK*}{UTF8}{gbsn}

\title{A Semantic Relevance Based Neural Network for Text Summarization and Text  Simplification}

\author{Shuming Ma\thanks{MOE Key Laboratory of Computational Linguistics, Peking University, Beijing, China and School of Electronics Engineering and Computer Science, Peking University, Beijing, China. E-mail: shumingma@pku.edu.cn.}}
\affil{Peking University}

\author{Xu Sun\thanks{MOE Key Laboratory of Computational Linguistics, Peking University, Beijing, China and School of Electronics Engineering and Computer Science, Peking University, Beijing, China. E-mail: xusun@pku.edu.cn.}}
\affil{Peking University}





\maketitle
\begin{abstract}
	Text summarization and text simplification are two major ways to simplify the text for poor readers, including children, non-native speakers, and the functionally illiterate. Text summarization is to produce a brief summary of the main ideas of the text, while text simplification aims to reduce the linguistic complexity of the text and retain the original meaning. Recently, most approaches for text summarization and text simplification are based on the sequence-to-sequence model, which achieves much success in many text generation tasks. However, although the generated simplified texts are similar to source texts literally, they have low semantic relevance. In this work, our goal is to improve semantic relevance between source texts and simplified texts for text summarization and text simplification. We 
	introduce a Semantic Relevance Based neural model to encourage high semantic similarity between texts and summaries. In our model, the source text is represented by a gated attention encoder, while the summary representation is produced by a decoder. Besides, the similarity score between the representations is maximized during training. Our experiments show that the proposed model outperforms the state-of-the-art systems on two benchmark corpus\footnote{Our code is available at https://github.com/shumingma/SRB}.
\end{abstract}

\section{Introduction}

Text summarization and text simplification is to make the text easier to read and understand, especially for poor readers, including children, non-native speakers, and the functionally illiterate. Text summarization is to simplify the texts at the document level. The source texts are often consist of many sentences or paragraphs, and the simplified texts are some brief sentences of the main ideas of the source texts. Text simplification is to simplify the texts at the sentence level. It aims to simplify the sentences to reduce the lexical and structure complexity. Unlike text summarization, it does not require the simplified sentences are shorter, but requires the words are simple to understand.

In some previous work, extractive summarization achieves satisfying performance by selecting a few sentences from source texts~\cite{extra04,discourse,extra15}. By extracting the sentences, the generated texts are grammatical, and retain the same meaning with the source texts. However, it does not simplify the texts but only shorten the texts. Some previous related work regards text simplification as a combination of three operations: splitting, deletion and paraphrasing, which requires some rule-based models or heavy syntactic features~\cite{ZhuEA2010,Woodsend2011,FilippovaEA2015}. 

Most recent approaches use sequence-to-sequence model for text summarization~\cite{abs,lcsts} and text simplification~\cite{NisioiEA2017,CaoEA2017,ZhangEA2017}. Sequence-to-sequence model is a widely used end-to-end framework for text generation, such as machine translation. It compresses the source text information into dense vectors with the neural encoder, and the neural decoder generates the target text using the compressed vectors.

For both text summarization and text simplification, the simplified texts must have high semantic relevance to the source texts. However, current sequence-to-sequence models tend to produce grammatical and coherent simplified texts regardless of the semantic relevance to source texts. Table~\ref{tab1} shows that the summary generated by a LSTM sequence-to-sequence model (Seq2seq) is similar to the source text literally, but it has low semantic relevance.

\begin{table}[tb]
	\centering
	\caption{An example of the simplified text generated by Seq2seq for summarization. The summary has high similarity to the text literally, but has low semantic relevance.
	}\label{tab1}
	\begin{tabular}{| l p{10.5cm}@{~} |}
		\hline
		Text & 昨晚，\textbf{中联航空}成都飞北京一架航班
		被发现有\textbf{多人}吸烟。后因天气原因，飞机
		备降太原\textbf{机场}。有乘客要求重新安检，机
		长决定继续飞行，引起机组人员与未吸烟
		乘客冲突。\\
		& Last night, \textbf{several people} were caught to smoke on a flight of \textbf{China United Airlines} from 
		Chendu to Beijing. Later the flight temporarily landed on Taiyuan \textbf{Airport}. Some passengers asked for a security check but were denied 
		by the captain, which led to a collision between crew and passengers. \\
		&\\
		Seq2seq & \textbf{中联航空机场}发生爆炸致\textbf{多人}死亡。\\
		& \textbf{China United Airlines} exploded in the \textbf{airport}, 
		leaving \textbf{several people} dead. \\
		&\\
		Gold & 航班多人吸烟机组人员与乘客冲突。 \\ 
		& Several people smoked on a flight which led
		to a collision between crew and passengers. \\
		\hline
	\end{tabular}
	\vspace{-0.1in}
\end{table}

In this work, our goal is to improve the semantic relevance between source texts and generated simplified texts for text summarization and text simplification. To achieve this goal, we propose a Semantic Relevance Based neural network model (SRB). In our model, we compress the source texts into dense vectors with the encoder, and decode the dense vector into simplified texts with the decoder. The encoder produces the representation of source texts, and the decoder produces the representation of the generated texts. A similarity evaluation component is introduced to measure the relevance of source texts and generated texts. During training, it maximizes the similarity score to encourage high semantic relevance between source texts and simplified texts. In order to better represent a long source text, we introduce a self-gated attention encoder to memory the input text. We conduct the experiments on three corpus, namely LCSTS, PWKP, and EW-SEW. Experiments show that our proposed model has better performance than the state-of-the-art systems on two benchmark corpus.

The contributions of this work are as follow:
\begin{itemize}
	\item We propose a Semantic Relevance Based neural network model (SRB) to improve the semantic relevance between source texts and generated simplified texts for text summarization and text simplification. A similarity evaluation component is introduced to measure the relevance of source texts and generated texts, and the similarity score is maximized to encourage high semantic relevance between source texts and simplified texts.
	\item We introduce a self-gated encoder to better represent a long redundant text. We perform the experiments on three corpus, namely LCSTS, PWKP, and EW-SEW. Experiments show that our proposed model outperforms the state-of-the-art systems on two benchmark corpus.
\end{itemize}

\section{Background: Sequence-to-sequence Model}

Most recent models for text summarization and text simplification are based on the sequence-to-sequence model. The sequence-to-sequence model is able to compress source texts $x=\{x_1,x_2,...,x_N\}$ into a continuous vector representation with an encoder, and then generates the simplified text $y=\{y_1,y_2,...,y_M\}$ with a decoder. In the previous work~\cite{NisioiEA2017,lcsts}, the encoder is a two layer Long Short-term Memory Network (LSTM)~\cite{LSTM}, which maps source texts into the hidden vector $\{h_1,h_2,...,h_N\}$. The decoder is a uni-directional LSTM, producing the hidden output $s_t$, which is the dense representation of the words at the $t^{th}$ time step. Finally, the word generator computes the distribution of output words $y_t$ with the hidden state $s_t$ and the parameter matrix $W$:
\begin{equation}\label{generator1}
p(y_t|x)=softmax{(Ws_t)}
\end{equation}

Attention mechanism is introduced to better capture context information of source texts~\cite{attention}. Attention vector $c_t$ is calculated by the weighted sum of encoder hidden states:
\begin{equation}\label{attention1}
c_t=\sum_{i=1}^{N}{\alpha_{ti}h_{i}}
\end{equation}
\begin{equation}\label{attention2}
\alpha_{ti}=\frac{e^{g(s_{t},h_{i})}}{\sum_{j=1}^{N}{e^{g(s_{t},h_{j})}}}
\end{equation}
where $g(s_{t},h_{i})$ is an attentive score between the decoder hidden state $s_t$ and the encoder hidden state $h_i$. When predicting an output word, the decoder takes account of the attention vector, which contains the alignment information between source texts and simplified texts. With the attention mechanism, the word generator computes the distribution of output words $y_t$:
\begin{equation}\label{generator2}
p(y_t|x)=softmax{(W\tilde{s_t})}
\end{equation}
\begin{equation}
\tilde{s_t}=\tanh(W_c[s_t;c_t])
\end{equation}

\begin{figure}[tb]
	\centering
	\begin{tabular}{@{}c@{}@{}c@{}@{}c@{}@{}c@{}}
		
		\epsfig{file=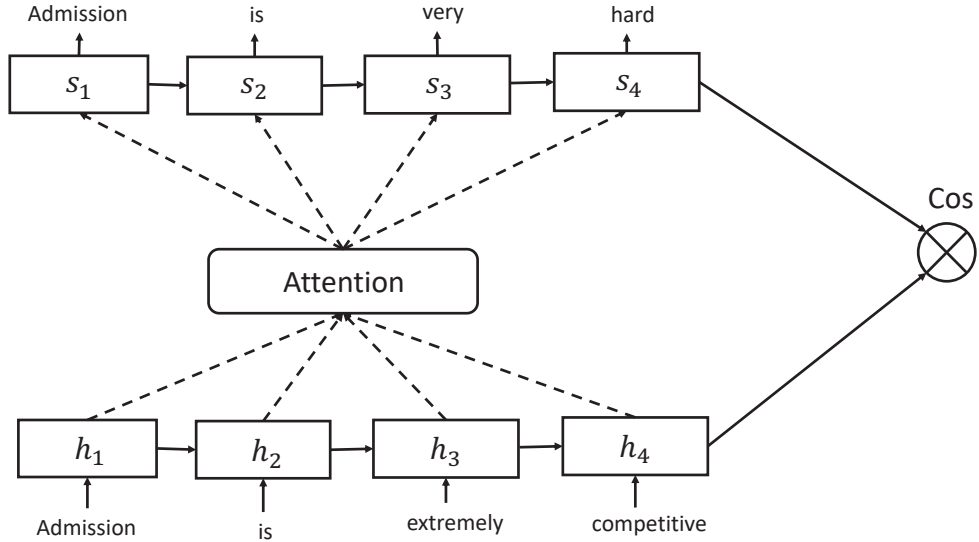,width=1.0\linewidth,clip=}
		
	\end{tabular}
	\caption{Our Semantic Relevance Based neural model. It consists of decoder (above), encoder (below) and cosine similarity function.
	}\label{fig1}
	\vspace{-0.1in}
\end{figure}

\section{Proposed Model}

Our goal is to improve the semantic relevance between source texts and simplified texts, so our proposed model encourages high similarity between their representations. Figure~\ref{fig1} shows our proposed model. The model consists of three components: encoder, decoder and a similarity function. The encoder compresses source texts into semantic vectors, and the decoder generates summaries and produces semantic vectors of the generated summaries. Finally, the similarity function evaluates the relevance between the sematic vectors of source texts and generated summaries. Our training objective is to maximize the similarity score so that the generated summaries have high semantic relevance to source texts.

\begin{figure}[tb]
	\centering
	\begin{tabular}{@{}c@{}@{}c@{}@{}c@{}@{}c@{}}
		
		\epsfig{file=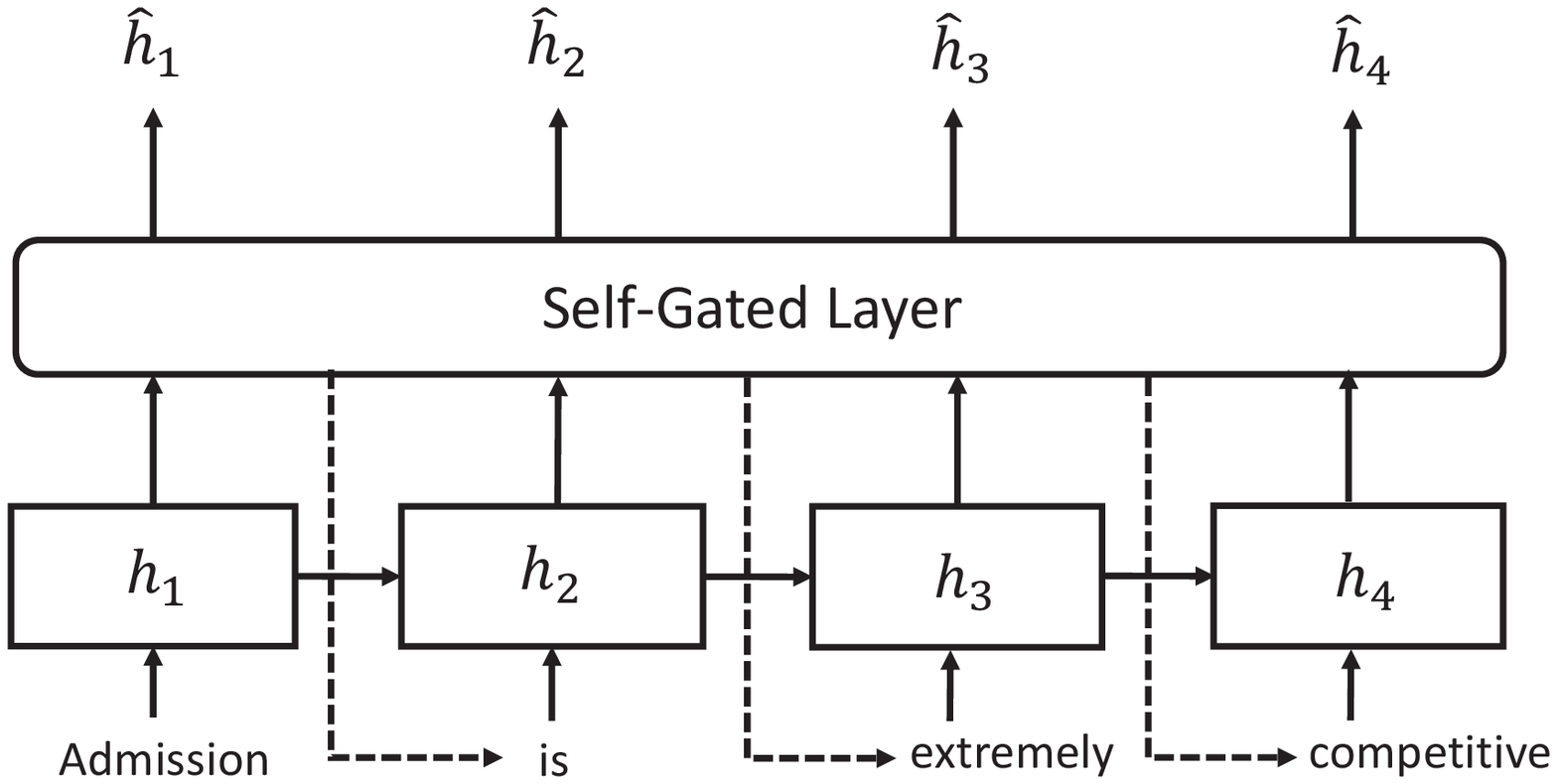,width=0.9\linewidth,clip=}
		
	\end{tabular}
	\caption{The self-gated encoder. It measure the importance of each word, and decide how much information is reserved as the representation of the texts.
	}\label{fig2}
	\vspace{-0.1in}
\end{figure}

\subsection{Self-gated Encoder}

The goal of the complex text encoder is to provide a series of dense representation of source texts for the decoder and the semantic relevance component. In the previous work~\cite{NisioiEA2017}, the complex text encoder is a two-layer uni-directional Long Short-term Memory Network (LSTM), which produces the dense representation $\{h_1,h_2,...,h_N\}$ from the source text $\{x_1,x_2,...,x_N\}$.

However, in text summarization and text simplification, source texts are usually very long and noisy. Therefore, some encoding information in the beginning of the texts will vanish until the end of the texts, which leads to bad representations of the texts. Bi-directional LSTM is an alternative to deal with the problem, but it needs double time to encoder the source texts, and it does not represents the middle of the texts well when the texts are too long. To solve the problem, we propose a self-gated encoder to better represent a long text. 

In text summarization and text simplification, some words or information in the source texts are unimportant, so they need to be simplified or discarded. Therefore, we introduce a self-gated encoder, which can reduce the unnecessary information and enhance the important information to represent a long text.

Self-gated encoder try to measure the importance of each word, and decide how much information is reserved as the representation of the texts. At each time step, every upcoming word $x_t$ is fed into the LSTM cell, which outputs the dense vector $h_t$:
\begin{equation}
h_t = f(x_t,h_{t-1})
\end{equation}
where $f$ is the LSTM function, and $h_t$ is the output vector of the LSTM cell. A feed-forward neural network is used to measure the importance and decide how much information is reversed:
\begin{equation}
\beta_t=sigmoid(g(h_t))
\end{equation}
where $g$ is the feed-forward neural network function, and $\beta_t$ measures the proportion of the reserved information. Finally, the reversed information is computed by multiplying $\beta_t$:
\begin{equation}
\hat{h}_t=\beta_th_t
\end{equation}
\begin{equation}
\hat{e}_{t+1}=\beta_te_{t+1}
\end{equation}
where $\hat{h}_t$ is the representation at the $t_{th}$ time step, and $\hat{e}_{t+1}$ is the input embedding of $x_{t+1}$ at the $t+1_{th}$ time step.


\subsection{Simplified Text Decoder}

The goal of the simplified text decoder is to generate a series of simplified words from the dense representation of source texts. In our model, the dense representations of the source texts are fed into an attention layer to generate the context vector $c_t$:
\begin{equation}\label{attention1}
c_t=\sum_{i=1}^{N}{\alpha_{ti}\hat{h}_{i}}
\end{equation}
\begin{equation}\label{attention2}
\alpha_{ti}=\frac{e^{g(s_{t},\hat{h}_{i})}}{\sum_{j=1}^{N}{e^{g(s_{t},\hat{h}_{j})}}}
\end{equation}
where $s_t$ is the dense representation of generated simplified computed by a two-layer LSTM. 

In this way, $c_t$ and $s_t$ respectively represent the context information of source texts and the target texts at the $t^{th}$ time step. To predict the $t^{th}$ word, the decoder uses $c_t$ and $s_t$ to generate the probability distribution of the candidate words:
\begin{equation}\label{generator2}
p_t(y|x)=softmax{(W\hat{s}_t)}
\end{equation}
\begin{equation}
\hat{s}_t=\tanh(W_c[s_t;c_t])
\end{equation}
where $W$ and $W_c$ are the parameter matrix of the output layer. Finally, the word with the highest probability is predicted:
\begin{equation}
y_t=argmax_{y'}{p_t(y'|x)}
\end{equation}

\subsection{Semantic Relevance}

Our goal is to compute the semantic relevance of source texts and generated texts given the source semantic vector $V_t$ and the generated sementic vector $V_s$. Here, we use cosine similarity to measure the semantic relevance, which is represented with a dot product and magnitude:
\begin{equation}
cos(V_s,V_t)=\frac{V_{s} \cdot V_{t}}{\Vert V_s \Vert \Vert V_t \Vert}
\end{equation}
Source texts and generated texts share the same language, so it is reasonable to assume that their semantic vectors are distributed in the same space. Cosine similarity is a good way to measure the distance between two vectors in the same space.

With the semantic relevance metric, the problem is how to get the semantic vector $V_s$ and $V_t$. There are several methods to represent a text or a sentence, such as mean pooling of LSTM output or reserving the last state of LSTM. In our model, we select the last state of the encoder as the representation of source texts:
\begin{equation}
V_s=\hat{h}_N
\end{equation}

A natural idea to get the semantic vector of a summary is to feed it into the encoder as well. However, this method wastes much time because we encode the same sentence twice. Actually, the last output of the decoder $\hat{s}_M$ contains information of both source text and generated summaries. We simply compute the semantic vector of the summary by subtracting $\hat{h}_N$ from $\hat{s}_M$:
\begin{equation}
V_{s}=\hat{s}_M-\hat{h}_N
\end{equation}
Previous work has proved that it is effective to represent a span of words without encoding them once more~\cite{wang16}.

\subsection{Training}

Given the model parameter $\theta$ and input text $x$, the model produces corresponding summary $y$ and semantic vector $V_s$ and $V_t$. The objective is to minimize the loss function:
\begin{equation}
L=-p(y|x;\theta)-\lambda cos(V_s,V_t)
\end{equation}
where $p(y|x;\theta)$ is the conditional probability of summaries given source texts, and is computed by the encoder-decoder model. $cos(V_s,V_t)$ is cosine similarity of semantic vectors $V_s$ and $V_t$. This term tries to maximize the semantic relevance between source input and target output.

We use Adam optimization method to train the model, with the default hyper-parameters: the learning rate $\alpha=0.001$, and $\beta_{1}=0.9$, $\beta_{2}=0.999$, $\epsilon=1e-8$.

\section{Experiments}

In this section, we present the evaluation of our model and show its performance on three popular corpus. Besides, we perform a case study to explain the semantic relevance between generated summary and source text.

\subsection{Datasets}

We introduce a Chinese text summarization dataset and two popular text simplification datasets. The simplification datasets are both from the alignments between English Wikipedia website\footnote{http://en.wikipedia.org} and Simple English Wikipedia website\footnote{http://simple.wikipedia.org}. The Simple English Wikipedia is built for ``the children and adults who are learning the English language'', and the articles are composed with ``easy words and short sentences''. Therefore, Simple English Wikipedia is a natural public simplified text corpus. Most of the text simplification benchmark datasets are constructed from Simple English Wikipedia.

\textbf{Large Scale Chinese Short Text Summarization Dataset (LCSTS).} LCSTS is constructed by \namecite{lcsts}. The dataset consists of more than 2.4 million text-summary pairs, constructed from a famous Chinese social media website called Sina Weibo\footnote{weibo.sina.com}. It is split into three parts, with 2,400,591 pairs in PART I, 10,666 pairs in PART II and 1,106 pairs in PART III. All the text-summary pairs in PART II and PART III are manually annotated with relevant scores ranged from 1 to 5, and we only reserve pairs with scores no less than 3. Following the previous work, we use PART I as training set, PART II as development set, and PART III as test set.

\textbf{Parallel Wikipedia Simplification Corpus (PWKP).} PWKP~\cite{ZhuEA2010} is a widely used benchmark for evaluating text simplification systems. It consists of aligned complex text from English WikiPedia (as of Aug. 22nd, 2009) and simple text from Simple Wikipedia (as of Aug. 17th, 2009). The dataset contains 108,016 sentence pairs, with 25.01 words on average per complex sentence and 20.87 words per simple sentence. Following the previous work~\cite{ZhangEA2017}, we remove the duplicate sentence pairs, and split the corpus with 89,042 pairs for training, 205 pairs for development and 100 pairs for test.

\textbf{English Wikipedia and Simple English Wikipedia (EW-SEW).} EW-SEW is a publicly available dataset provided by \namecite{HwangEA2015}. To build the corpus, they first align the complex-simple sentence pairs, score the semantic similarity between the complex sentence and the simple sentence, and classify each sentence pair as a good, good partial, partial, or bad match. Following the previous work~\cite{NisioiEA2017}, we discard the unclassified matches, and use the good matches and partial matches with a scaled threshold greater than 0.45. The corpus contains about 150K good matches and 130K good partial matches. We use this corpus as the training set, and the dataset provided by Xu et al.~\cite{XuEA2016} as the development set and the test set. The development set consists of 2,000 sentence pairs, and the test set contains 359 sentence pairs. Besides, each complex sentence is paired with 8 reference simplified sentences provided by Amazon Mechanical Turk workers.

\subsection{Settings}

We describe the experimental details of text summarization and text simplification respectively.

\noindent\textbf{Text Summarization.} To alleviate the risk of word segmentation mistakes~\cite{Xu2016Dependency,SunEA2012}, we use Chinese character sequences as both source inputs and target outputs. We limit the model vocabulary size to 4000, which covers most of the common characters. Each character is represented by a random initialized word embedding. We tune our parameter on the development set. In our model, the embedding size is 400, the hidden state size of encoder-decoder is 500, and the size of gated attention network is 1000. We use Adam optimizer to learn the model parameters, and the batch size is set as 32. The parameter $\lambda$ is 0.0001. Both the encoder and decoder are based on LSTM unit. Following the previous work~\cite{lcsts}, our evaluation metric is F-score of ROUGE: ROUGE-1, ROUGE-2 and ROUGE-L~\cite{rough}.

\noindent\textbf{Text Simplification.} The text simplification datasets contain a lot of named entities, which makes the vocabulary too large. To reduce the vocabulary size, we follow the setting by \namecite{ZhangEA2017}. We recognize the named entities with the Stanford CoreNLP tagger~\cite{ManningEA2014}, and replace the named entities with the anonymous symbols \emph{NE@N}, where \emph{NE}$\in$\{PER, LOC, ORG, MISC\} where $N$ represents the $N^{th}$ entity in the sentence. To limit the vocabulary size, we prune the vocabulary to top 50,000 most frequent words, and replace the rest words with the UNK symbols. At test time, we replace the UNK symbols with the highest probability score from the attention alignment matrix following Jean et al.~\cite{mapattention}. We filter out sentence pairs whose lengths exceed 100 words in the training set. The encoder is implemented on LSTM, and the decoder is based on LSTM with Luong style attention~\cite{stanfordattention}. We tune our hyper-parameter on the development set. The model has two LSTM layers. The hidden size of LSTM is 256, and the embedding size is 256. We use Adam optimizer~\cite{KingmaBa2014} to learn the parameters, and the batch size is set to be 64. We set the dropout rate~\cite{dropout} to be 0.4. All of the gradients are clipped when the norm exceeds 5. The evaluation metric is BLEU score.

\subsection{Baseline Systems}

\noindent\textbf{Seq2seq}. We first compare our model with a basic sequence-to-sequence model~\cite{seq2seq}. It is a widely used model to generate texts, so it is an important baseline.

\noindent\textbf{Seq2seq-Attention}. Seq2seq-Attention~\cite{attention} is a sequence-to-sequence framework with neural attention. Attention mechanism helps capture the context information of source texts. This model is a stronger baseline system.

\subsection{Results}

\begin{table}[tb]
	\centering
	\caption{Results of our model and baseline systems. Our models achieve substantial improvement of all ROUGE scores over baseline systems. The results are reported on the test sets.(W: Word level; C: Character level).} \label{tab2}
	\newcommand{\tabincell}[2]{\begin{tabular}{@{}#1@{}}#2\end{tabular}}
	\begin{tabular}{|l|c|c|c|}
		\hline
		Model  & ROUGE-1 & ROUGE-2 & ROUGE-L \\
		\hline
		Seq2seq (W)~\cite{lcsts}  &     17.7  & 8.5  &  15.8 \\
		Seq2seq (C)~\cite{lcsts}  &       21.5  & 8.9  &  18.6  \\
		Seq2seq-Attention (W)~\cite{lcsts} & 26.8  & 16.1  &  24.1  \\
		Seq2seq-Attention (C)~\cite{lcsts} & 29.9 & 17.4  &  27.2  \\
		\hline
		Seq2seq-Attention (C) (our implementation) & 30.1  & 17.9  & 27.2  \\
		\textbf{SRB (C) (our proposal)} & \textbf{33.3}   & \textbf{20.0}  & \textbf{30.1} \\
		\hline
	\end{tabular}
\end{table}

\begin{table}[tb]
	\centering
	\caption{Comparison with our model and the recent neural models for text simplification. Our models achieve substantial improvement of BLEU score over baseline systems. The results are reported on the test sets.}\label{tab3}
	\begin{tabular}{|@{~} l@{~}|c|}
		\hline
		PWKP & BLEU \\ 
		\hline
		Seq2seq-Attention~\cite{NisioiEA2017} & 47.52\\
		Seq2seq-Attention-w2v~\cite{NisioiEA2017} &  48.10 \\ 
		\hline
		Seq2seq-Attention (our implementation) & 48.26 \\ 
		\textbf{SRB (our proposal)}  & \textbf{50.18}  \\ 
		\hline
		\hline
		EW-SEW & BLEU  \\ 
		\hline
		Seq2seq-Attention~\cite{NisioiEA2017} &  84.70  \\
		Seq2seq-Attention-w2v~\cite{NisioiEA2017}  & 87.50  \\ 
		\hline
		Seq2seq-Attention (our implementation)  & 88.97  \\ 
		\textbf{SRB (our proposal)} &  \textbf{89.84}  \\ 
		\hline
	\end{tabular}
\end{table}

\begin{table}[tb]
	\centering
	\caption{Results of our model and state-of-the-art systems. COPYNET incorporates copying mechanism to solve out-of-vocabulary problem, so its has higher ROUGE scores. Our model does not incorporate this mechanism
		currently. In the future work, we will implement this technique to further improve the performance. The results are reported on the test sets. (Word: Word level; Char: Character level; R-1: F-score of ROUGE-1; R-2: F-score of ROUGE-2; R-L: F-score of ROUGE-L)} \label{tab4}
	\newcommand{\tabincell}[2]{\begin{tabular}{@{}#1@{}}#2\end{tabular}}
	\begin{tabular}{|l|c|c|c|}
		\hline
		Model & ROUGE-1 & ROUGE-2 & ROUGE-L \\
		\hline
		Seq2seq (W)~\cite{lcsts} & 26.8 & 16.1 &  24.1 \\
		Seq2seq (C)~\cite{lcsts} & 29.9 & 17.4  &  27.2 \\
		Seq2seq-Attention (W)~\cite{lcsts} & 26.8 & 16.1 &  24.1 \\
		Seq2seq-Attention (C)~\cite{lcsts} & 29.9 & 17.4  &  27.2 \\
		COPYNET (C)~\cite{copynet} & \textbf{35.0} & \textbf{22.3} & \textbf{32.0} \\
		\hline
		\textbf{SRB (C) (our proposal)} & 33.3 & 20.0 & 30.1 \\
		\hline
	\end{tabular}
\end{table}

\begin{table}[tb]
	\centering
	\caption{Results of our model and state-of-the-art systems. SRB achieves the best BLEU scores compared with the related systems on PWKP and EW-SEW. The results are reported on the test sets.} \label{tab5}
	\newcommand{\tabincell}[2]{\begin{tabular}{@{}#1@{}}#2\end{tabular}}
	\begin{tabular}{|l|c|}
		\hline
		PWKP & BLEU \\
		\hline
		NTS~\cite{NisioiEA2017} & 47.52  \\
		NTS-w2v~\cite{NisioiEA2017} & 48.10  \\
		DRESS~\cite{ZhangEA2017} &  34.53  \\
		DRESS-LS~\cite{ZhangEA2017} &  36.32   \\ 
		\hline
		\textbf{SRB (our proposal)} &  \textbf{50.18}  \\ \hline
		\multicolumn{2}{c}{} \\
		\hline
		EW-SEW & BLEU  \\
		\hline
		PBMT-R (Wubben et al. 2012) &  67.79  \\
		SBMT-SARI~\cite{XuEA2016} &  73.62   \\ \hline
		NTS~\cite{NisioiEA2017} &  84.70  \\
		NTS-w2v~\cite{NisioiEA2017} & 87.50  \\
		DRESS~\cite{ZhangEA2017} &  77.18   \\
		DRESS-LS~\cite{ZhangEA2017} &  80.12   \\
		\hline
		\textbf{SRB (our proposal)} &  \textbf{89.84}    \\ \hline
	\end{tabular}
\end{table}

\begin{table}[tb]
	\centering
	\caption{An Example of SRB generated summary on LCSTS dataset, compared with the system output of Seq2seq-Attention and the reference.
	}\label{tab6}
	\begin{tabular}{|l p{10.5cm}@{~} |}
		\hline
		Text & 仔细一算，上海的互联网公司不乏成功
		案例，但最终成为BAT一类巨头的几乎没有,
		这也能解释为何纳税百强的榜单中鲜少互联
		网公司的身影。有一类是被并购，比如：易
		趣、土豆网、PPS、PPTV、一号店等；有一
		类是数年偏安于细分市场。\\
		& With careful calculation, there are many successful Internet companies in Shanghai, but few
		of them becomes giant company like BAT. This is also the reason why few Internet companies are listed in top hundred companies of paying tax. Some of them are merged, such as Ebay, Tudou, PPS, PPTV, Yihaodian and so on.
		Others are satisfied with segment market for years.\\
		&\\
		Reference &为什么上海出不了互联网巨头？\\
		&Why Shanghai comes out no giant company?\\
		&\\
		Seq2seq-A&上海的互联网巨头。\\
		&Shanghai's giant company.\\
		&\\
		SRB&上海鲜少互联网巨头的身影。\\
		&Shanghai has few giant companies.\\
		\hline
	\end{tabular}
	\vspace{-0.01in}
\end{table}

\begin{table}[tb]
	\centering
	\caption{An examples of different text simplification system outputs in EW-SEW dataset. Differences from the source texts are shown in bold.
	}\label{tab7}
	\begin{tabular}{| l  p{10.5cm}@{~} |}
		\hline
		Source & Depending on the context, another closely-related meaning of constituent is that of a citizen residing in the area governed, represented, or otherwise served by a politician; sometimes this is restricted to citizens who elected the politician. \\ 
		Reference & \textbf{The word constituent can also be used to refer to} a citizen \textbf{who lives} in the area that is governed, represented, or otherwise served by a politician; sometimes \textbf{the word} is restricted to citizens who elected the politician. \\
		NTS & Depending on the context, another closely-related meaning of constituent is that of a citizen \textbf{living} in the area governed, represented, or otherwise served by a politician; sometimes this is restricted to citizens who elected the politician. \\
		NTS-w2v & This is restricted to citizens who elected the politician. \\
		PBMT-R & Depending on the context and meaning of closely-related \textbf{siemens-martin -rrb- is a} citizen living in the area, or otherwise, \textbf{was governed by a 1924-1930 shurba}; this is restricted to people who elected \textbf{it}. \\
		SBMT-SARI & \textbf{In terms of} the context, another closely-related \textbf{sense of the component} is that of a citizen \textbf{living} in the area \textbf{covered, make up, or if not, served by a policy}; sometimes this is \textbf{limited} to the people who elected the \textbf{policy}.
		\\
		SRB & Depending on the context, another closely-related meaning of constituent is that of a citizen \textbf{living} in the area governed, represented, or otherwise served by a politician; sometimes \textbf{the word} is restricted to citizens who elected the politician. \\
		\hline
	\end{tabular}
	\vspace{-0.1in}
\end{table}

We compare our model with above baseline systems, including Seq2seq and Seq2seq-Attention. We refer to our proposed Semantic Relevance Based neural model as \textbf{SRB}. Table~\ref{tab2} shows the results of our models and baseline systems on LCSTS. As shown in Table~\ref{tab2}, the models at the character level achieve better performance than the models at the word level. Therefore, we implement our model at the character level. For fair comparison, we also implement a Seq2seq-Attention model following the details in the previous work~\cite{lcsts}. Our implementation of Seq2seq-Attention has better score, mainly because we tune the hyper-parameters well on the development set.
We can see SRB outperforms both Seq2seq and Seq2seq-Attention with the F-score of 33.3 ROUGE-1, 20.0 ROUGE-2 and 30.1 ROUGE-L. 

Table~\ref{tab3} shows the results in two text simplification corpus. We compare our model with Seq2seq-Attention and Seq2seq-Attention-w2v. Seq2seq-Attention-w2v is a Seq2seq-Attention with pretrain word embeddings. We also implement a Seq2seq-Attention model, and carefully tune it on the development set. Our implementation get 48.26 BLEU score on PWKP, and 88.97 BLEU score on EW-SEW. Our SRB outperforms all of the baseline systems, with the BLEU score of 50.18 on PWKP and the BLEU score of 89.84 on EW-SEW. 

Table~\ref{tab4} summarizes the results of our model and state-of-the-art systems. COPYNET has the highest scores, because it incorporates copying mechanism to deals with out-of-vocabulary word problem.
In this paper, we do not implement this mechanism in our model.
Our model can also be improved with these additional
techniques, which, however, are not the focus of this
paper.

We also compare SRB with other models for text simplification, which are not limit to neural models. Table~\ref{tab5} summarizes the results of SRB and the related systems. On PWKP dataset, we compare SRB with NTS, NTS-w2v, DRESS and DRESS-LS. We run the public release code of NTS and NTS-w2v provided by \namecite{NisioiEA2017}, and get the BLEU score of 47.52 and 48.10 respectively. As for DRESS and DRESS-LS, we use the scores reported by \namecite{ZhangEA2017}. 
The goal of DRESS is not to generate the outputs closer to the references, so BLEU of DRESS and DRESS-LS are relatively lower than NTS and NTS-w2v. SRB achieves a BLEU score of 50.18, outperforming all of the previous systems. On EW-SEW dataset, we compare WEAN-dot with PBMT-R, SBMT-SARI, and the neural models described above. We do not find any public release code of PBMT-R and SBMT-SARI. Fortunately, \namecite{XuEA2016} provides the predictions of PBMT-R and SBMT-SARI on EW-SEW test set, so that we can compare our model with these systems. It shows that the neural models have better performance in BLEU, and WEAN-dot achieves the best BLEU score with 89.84.

\subsection{Case Study}

Table~\ref{tab6} is an example to show the semantic relevance between the source text and the summary. It shows that the main idea of the source text is
about the reason why Shanghai has few giant company. RNN context produces ``Shanghai's giant companies'' which is literally similar to the source text, while
SRB generates ``Shanghai has few giant companies'', which is closer to the main idea in semantics. It concludes that SRB produces summaries with higher
semantic similarity to texts.

Table~\ref{tab7} shows an examples of different text simplification system outputs on EW-SEW. NTS-w2v omits so many words that it lacks a lot of information. PBMT-R generates some irrelevant words, like 'siemens-martin', '-rrb-', and '-shurba', which hurts the fluency and adequacy of the generated sentence. SBMT-SARI is able to generate a fluent sentence, but the meaning is different from the source text, and even more difficult to understand.
Compared with the statistic model, SRB generates a more fluent sentence. Besides, SRB improves the semantic revelance between the source texts and the generated texts, so the generated sentence is semantically correct, and very close to the original meaning.

\section{Related Work}

Abstractive text summarization has achieved successful performance thanks to the sequence-to-sequence model~\cite{seq2seq} and attention mechanism~\cite{attention}.
\namecite{abs} first used an attention-based encoder to compress texts and a neural network language decoder to generate summaries.
Following this work, recurrent encoder was introduced to text summarization, and gained better performance~\cite{rnnheadline,ras}. Towards Chinese texts, \namecite{lcsts}
built a large corpus of Chinese short text summarization. To deal with unknown word problem, \namecite{ibmsummarization} proposed a generator-pointer model
so that the decoder is able to generate words in source texts. \namecite{copynet} also solved this issue by incorporating copying mechanism.
Besides, \namecite{minimum} proposes a minimum risk training method which optimizes the parameters with the target of rouge scores.

\namecite{ZhuEA2010} constructs a wikipedia dataset, and proposes a tree-based simplification model, which is the first statistical simplification model covering splitting, dropping, reordering and substitution integrally. \namecite{Woodsend2011} introduces a data-driven model based on quasi-synchronous grammar, which captures structural mismatches and complex rewrite operations. \namecite{WubbenEA2012} presents a method for text simplification using phrase based machine translation with re-ranking the outputs. \namecite{Kauchak2013} proposes a text simplification corpus, and evaluates language modeling for text simplification on the proposed corpus.

\namecite{NarayanEA2014} propose a hybrid approach to sentence simplification which combines deep semantics and monolingual machine translation. \namecite{HwangEA2015} introduces a parallel simplification corpus by evaluating the similarity between the source text and the simplified text based on WordNet. \namecite{LigthLS} propose an unsupervised approach to lexical simplification that makes use of word vectors and require only regular corpora. \namecite{XuEA2016} design automatic metrics for text simplification, and they introduce a statistic machine translation, and tune with the proposed automatic metrics. 

Recently, most works focus on the neural sequence-to-sequence model. \namecite{NisioiEA2017} present a sequence-to-sequence model, and re-ranks the predictions with BLEU and SARI. \namecite{ZhangEA2017} propose a deep reinforcement learning model to improve the simplicity, fluency and adequacy of the simplified texts. \namecite{CaoEA2017} introduce a novel sequence-to-sequence model to join copying and restricted generation for text simplification.

Our work is also related to the encoder-decoder framework~\cite{ChoEA2014} and the attention mechanism~\cite{attention}. Encoder-decoder framework, like sequence-to-sequence model, has achieved success in machine translation~\cite{seq2seq,mapattention,stanfordattention}, text summarization~\cite{abs,ras,ibmsummarization,CaoEA2016}, and other natural language processing tasks. Neural attention model is first proposed by \namecite{attention}. There are many other methods to improve neural attention model~\cite{mapattention,stanfordattention}.

\section{Conclusion}

In this work, our goal is to improve the semantic relevance between source texts and generated simplified texts for text summarization and text simplification. To achieve this goal, we propose a Semantic Relevance Based neural network model (SRB). A similarity evaluation component is introduced to measure the relevance of source texts and generated texts. During training, it maximizes the similarity score to encourage high semantic relevance between source texts and simplified texts. In order to better represent a long source text, we introduce a self-gated attention encoder to memory the input text. We conduct the experiments on three corpus, namely LCSTS, PWKP, and EW-SEW. Experiments show that our proposed model has better performance than the state-of-the-art systems on the benchmark corpus.

\section*{Acknowledgements}

This work was supported in part by National Natural Science Foundation of China (No. 61673028), and an Okawa Research Grant (2016). Xu Sun is the corresponding author of this paper. This work is a substantial extension of the conference version presented at ACL 2017~\cite{srb}.

\starttwocolumn
\bibliography{compling_style}

\begin{thebibliography}{37}
\expandafter\ifx\csname natexlab\endcsname\relax\def\natexlab#1{#1}\fi

\bibitem[{Ayana et~al.(2016)Ayana, Shen, Liu, and Sun}]{minimum}
Ayana, Shiqi Shen, Zhiyuan Liu, and Maosong Sun. 2016.
\newblock Neural headline generation with minimum risk training.
\newblock \emph{CoRR}, abs/1604.01904.

\bibitem[{Bahdanau, Cho, and Bengio(2014)}]{attention}
Bahdanau, Dzmitry, Kyunghyun Cho, and Yoshua Bengio. 2014.
\newblock Neural machine translation by jointly learning to align and
  translate.
\newblock \emph{CoRR}, abs/1409.0473.

\bibitem[{Cao et~al.(2016)Cao, Li, Li, Wei, and Li}]{CaoEA2016}
Cao, Ziqiang, Wenjie Li, Sujian Li, Furu Wei, and Yanran Li. 2016.
\newblock Attsum: Joint learning of focusing and summarization with neural
  attention.
\newblock In \emph{{COLING} 2016, 26th International Conference on
  Computational Linguistics, Proceedings of the Conference: Technical Papers,
  December 11-16, 2016, Osaka, Japan}, pages 547--556.

\bibitem[{Cao et~al.(2017)Cao, Luo, Li, and Li}]{CaoEA2017}
Cao, Ziqiang, Chuwei Luo, Wenjie Li, and Sujian Li. 2017.
\newblock Joint copying and restricted generation for paraphrase.
\newblock In \emph{Proceedings of the Thirty-First {AAAI} Conference on
  Artificial Intelligence}, pages 3152--3158.

\bibitem[{Cao et~al.(2015)Cao, Wei, Li, Li, Zhou, and Wang}]{extra15}
Cao, Ziqiang, Furu Wei, Sujian Li, Wenjie Li, Ming Zhou, and Houfeng Wang.
  2015.
\newblock Learning summary prior representation for extractive summarization.
\newblock In \emph{Proceedings of the 53rd Annual Meeting of the Association
  for Computational Linguistics and the 7th International Joint Conference on
  Natural Language Processing of the Asian Federation of Natural Language
  Processing, {ACL} 2015, July 26-31, 2015, Beijing, China, Volume 2: Short
  Papers}, pages 829--833.

\bibitem[{Cheng and Lapata(2016)}]{discourse}
Cheng, Jianpeng and Mirella Lapata. 2016.
\newblock Neural summarization by extracting sentences and words.
\newblock In \emph{Proceedings of the 54th Annual Meeting of the Association
  for Computational Linguistics, {ACL} 2016, August 7-12, 2016, Berlin,
  Germany, Volume 1: Long Papers}.

\bibitem[{Cho et~al.(2014)Cho, van Merrienboer, G{\"{u}}l{\c{c}}ehre, Bahdanau,
  Bougares, Schwenk, and Bengio}]{ChoEA2014}
Cho, Kyunghyun, Bart van Merrienboer, {\c{C}}aglar G{\"{u}}l{\c{c}}ehre,
  Dzmitry Bahdanau, Fethi Bougares, Holger Schwenk, and Yoshua Bengio. 2014.
\newblock Learning phrase representations using {RNN} encoder-decoder for
  statistical machine translation.
\newblock In \emph{Proceedings of the 2014 Conference on Empirical Methods in
  Natural Language Processing, {EMNLP} 2014}, pages 1724--1734.

\bibitem[{Chopra, Auli, and Rush(2016)}]{ras}
Chopra, Sumit, Michael Auli, and Alexander~M. Rush. 2016.
\newblock Abstractive sentence summarization with attentive recurrent neural
  networks.
\newblock In \emph{{NAACL} {HLT} 2016, The 2016 Conference of the North
  American Chapter of the Association for Computational Linguistics: Human
  Language Technologies}, pages 93--98.

\bibitem[{Filippova et~al.(2015)Filippova, Alfonseca, Colmenares, Kaiser, and
  Vinyals}]{FilippovaEA2015}
Filippova, Katja, Enrique Alfonseca, Carlos~A. Colmenares, Lukasz Kaiser, and
  Oriol Vinyals. 2015.
\newblock Sentence compression by deletion with lstms.
\newblock In \emph{Proceedings of the 2015 Conference on Empirical Methods in
  Natural Language Processing, {EMNLP}}, pages 360--368.

\bibitem[{Glava\v{s} and \v{S}tajner(2015)}]{LigthLS}
Glava\v{s}, Goran and Sanja \v{S}tajner. 2015.
\newblock Simplifying lexical simplification: Do we need simplified corpora?
\newblock In \emph{Proceedings of the 53rd Annual Meeting of the Association
  for Computational Linguistics, {ACL}}, pages 63--68.

\bibitem[{Gu et~al.(2016)Gu, Lu, Li, and Li}]{copynet}
Gu, Jiatao, Zhengdong Lu, Hang Li, and Victor O.~K. Li. 2016.
\newblock Incorporating copying mechanism in sequence-to-sequence learning.
\newblock In \emph{Proceedings of the 54th Annual Meeting of the Association
  for Computational Linguistics, {ACL} 2016}.

\bibitem[{Hochreiter and Schmidhuber(1997)}]{LSTM}
Hochreiter, Sepp and J{\"{u}}rgen Schmidhuber. 1997.
\newblock Long short-term memory.
\newblock \emph{Neural Computation}, 9(8):1735--1780.

\bibitem[{Hu, Chen, and Zhu(2015)}]{lcsts}
Hu, Baotian, Qingcai Chen, and Fangze Zhu. 2015.
\newblock {LCSTS:} {A} large scale chinese short text summarization dataset.
\newblock In \emph{Proceedings of the 2015 Conference on Empirical Methods in
  Natural Language Processing, {EMNLP} 2015, Lisbon, Portugal, September 17-21,
  2015}, pages 1967--1972.

\bibitem[{Hwang et~al.(2015)Hwang, Hajishirzi, Ostendorf, and Wu}]{HwangEA2015}
Hwang, William, Hannaneh Hajishirzi, Mari Ostendorf, and Wei Wu. 2015.
\newblock Aligning sentences from standard wikipedia to simple wikipedia.
\newblock In \emph{{NAACL} {HLT} 2015}, pages 211--217.

\bibitem[{Jean et~al.(2015)Jean, Cho, Memisevic, and Bengio}]{mapattention}
Jean, S{\'{e}}bastien, KyungHyun Cho, Roland Memisevic, and Yoshua Bengio.
  2015.
\newblock On using very large target vocabulary for neural machine translation.
\newblock In \emph{Proceedings of the 53rd Annual Meeting of the Association
  for Computational Linguistics, {ACL} 2015}, pages 1--10.

\bibitem[{Kauchak(2013)}]{Kauchak2013}
Kauchak, David. 2013.
\newblock Improving text simplification language modeling using unsimplified
  text data.
\newblock In \emph{Proceedings of the 51st Annual Meeting of the Association
  for Computational Linguistics, {ACL}}, pages 1537--1546.

\bibitem[{Kingma and Ba(2014)}]{KingmaBa2014}
Kingma, Diederik~P. and Jimmy Ba. 2014.
\newblock Adam: {A} method for stochastic optimization.
\newblock \emph{CoRR}, abs/1412.6980.

\bibitem[{Lin and Hovy(2003)}]{rough}
Lin, Chin{-}Yew and Eduard~H. Hovy. 2003.
\newblock Automatic evaluation of summaries using n-gram co-occurrence
  statistics.
\newblock In \emph{Human Language Technology Conference of the North American
  Chapter of the Association for Computational Linguistics, {HLT-NAACL} 2003}.

\bibitem[{Lopyrev(2015)}]{rnnheadline}
Lopyrev, Konstantin. 2015.
\newblock Generating news headlines with recurrent neural networks.
\newblock \emph{CoRR}, abs/1512.01712.

\bibitem[{Luong, Pham, and Manning(2015)}]{stanfordattention}
Luong, Thang, Hieu Pham, and Christopher~D. Manning. 2015.
\newblock Effective approaches to attention-based neural machine translation.
\newblock In \emph{Proceedings of the 2015 Conference on Empirical Methods in
  Natural Language Processing, {EMNLP} 2015}, pages 1412--1421.

\bibitem[{Ma et~al.(2017)Ma, Sun, Xu, Wang, Li, and Su}]{srb}
Ma, Shuming, Xu~Sun, Jingjing Xu, Houfeng Wang, Wenjie Li, and Qi~Su. 2017.
\newblock Improving semantic relevance for sequence-to-sequence learning of
  chinese social media text summarization.
\newblock In \emph{Proceedings of the 55th Annual Meeting of the Association
  for Computational Linguistics, {ACL} 2017, Vancouver, Canada, July 30 -
  August 4, Volume 2: Short Papers}, pages 635--640.

\bibitem[{Manning et~al.(2014)Manning, Surdeanu, Bauer, Finkel, Bethard, and
  McClosky}]{ManningEA2014}
Manning, Christopher~D., Mihai Surdeanu, John Bauer, Jenny~Rose Finkel, Steven
  Bethard, and David McClosky. 2014.
\newblock The stanford corenlp natural language processing toolkit.
\newblock In \emph{Proceedings of the 52nd Annual Meeting of the Association
  for Computational Linguistics, {ACL}}, pages 55--60.

\bibitem[{Nallapati et~al.(2016)Nallapati, Zhou, dos Santos,
  G{\"{u}}l{\c{c}}ehre, and Xiang}]{ibmsummarization}
Nallapati, Ramesh, Bowen Zhou, C{\'{\i}}cero~Nogueira dos Santos, {\c{C}}aglar
  G{\"{u}}l{\c{c}}ehre, and Bing Xiang. 2016.
\newblock Abstractive text summarization using sequence-to-sequence rnns and
  beyond.
\newblock In \emph{Proceedings of the 20th {SIGNLL} Conference on Computational
  Natural Language Learning, CoNLL 2016, Berlin, Germany, August 11-12, 2016},
  pages 280--290.

\bibitem[{Narayan and Gardent(2014)}]{NarayanEA2014}
Narayan, Shashi and Claire Gardent. 2014.
\newblock Hybrid simplification using deep semantics and machine translation.
\newblock In \emph{Proceedings of the 52nd Annual Meeting of the Association
  for Computational Linguistics, {ACL}}, pages 435--445.

\bibitem[{Nisioi et~al.(2017)Nisioi, Stajner, Ponzetto, and
  Dinu}]{NisioiEA2017}
Nisioi, Sergiu, Sanja Stajner, Simone~Paolo Ponzetto, and Liviu~P. Dinu. 2017.
\newblock Exploring neural text simplification models.
\newblock In \emph{Proceedings of the 55th Annual Meeting of the Association
  for Computational Linguistics, {ACL}}, pages 85--91.

\bibitem[{Radev et~al.(2004)Radev, Allison, Blair{-}Goldensohn, Blitzer,
  {\c{C}}elebi, Dimitrov, Dr{\'{a}}bek, Hakim, Lam, Liu, Otterbacher, Qi,
  Saggion, Teufel, Topper, Winkel, and Zhang}]{extra04}
Radev, Dragomir~R., Timothy Allison, Sasha Blair{-}Goldensohn, John Blitzer,
  Arda {\c{C}}elebi, Stanko Dimitrov, Elliott Dr{\'{a}}bek, Ali Hakim, Wai Lam,
  Danyu Liu, Jahna Otterbacher, Hong Qi, Horacio Saggion, Simone Teufel,
  Michael Topper, Adam Winkel, and Zhu Zhang. 2004.
\newblock {MEAD} - {A} platform for multidocument multilingual text
  summarization.
\newblock In \emph{Proceedings of the Fourth International Conference on
  Language Resources and Evaluation, {LREC} 2004}.

\bibitem[{Rush, Chopra, and Weston(2015)}]{abs}
Rush, Alexander~M., Sumit Chopra, and Jason Weston. 2015.
\newblock A neural attention model for abstractive sentence summarization.
\newblock In \emph{Proceedings of the 2015 Conference on Empirical Methods in
  Natural Language Processing, {EMNLP} 2015, Lisbon, Portugal, September 17-21,
  2015}, pages 379--389.

\bibitem[{Srivastava et~al.(2014)Srivastava, Hinton, Krizhevsky, Sutskever, and
  Salakhutdinov}]{dropout}
Srivastava, Nitish, Geoffrey~E. Hinton, Alex Krizhevsky, Ilya Sutskever, and
  Ruslan Salakhutdinov. 2014.
\newblock Dropout: a simple way to prevent neural networks from overfitting.
\newblock \emph{Journal of Machine Learning Research}, 15(1):1929--1958.

\bibitem[{Sun, Wang, and Li(2012)}]{SunEA2012}
Sun, Xu, Houfeng Wang, and Wenjie Li. 2012.
\newblock Fast online training with frequency-adaptive learning rates for
  chinese word segmentation and new word detection.
\newblock In \emph{Proceedings of ACL'12}, pages 253--262.

\bibitem[{Sutskever, Vinyals, and Le(2014)}]{seq2seq}
Sutskever, Ilya, Oriol Vinyals, and Quoc~V. Le. 2014.
\newblock Sequence to sequence learning with neural networks.
\newblock In \emph{Advances in Neural Information Processing Systems 27: Annual
  Conference on Neural Information Processing Systems 2014}, pages 3104--3112.

\bibitem[{Wang and Chang(2016)}]{wang16}
Wang, Wenhui and Baobao Chang. 2016.
\newblock Graph-based dependency parsing with bidirectional {LSTM}.
\newblock In \emph{Proceedings of the 54th Annual Meeting of the Association
  for Computational Linguistics, {ACL}}.

\bibitem[{Woodsend and Lapata(2011)}]{Woodsend2011}
Woodsend, Kristian and Mirella Lapata. 2011.
\newblock Learning to simplify sentences with quasi-synchronous grammar and
  integer programming.
\newblock In \emph{Proceedings of the 2011 Conference on Empirical Methods in
  Natural Language Processing, {EMNLP}}, pages 409--420.

\bibitem[{Wubben, van~den Bosch, and Krahmer(2012)}]{WubbenEA2012}
Wubben, Sander, Antal van~den Bosch, and Emiel Krahmer. 2012.
\newblock Sentence simplification by monolingual machine translation.
\newblock In \emph{The 50th Annual Meeting of the Association for Computational
  Linguistics, Proceedings of the Conference}, pages 1015--1024.

\bibitem[{Xu and Sun(2016)}]{Xu2016Dependency}
Xu, Jingjing and Xu~Sun. 2016.
\newblock Dependency-based gated recursive neural network for chinese word
  segmentation.
\newblock In \emph{Meeting of the Association for Computational Linguistics},
  pages 567--572.

\bibitem[{Xu et~al.(2016)Xu, Napoles, Pavlick, Chen, and
  Callison{-}Burch}]{XuEA2016}
Xu, Wei, Courtney Napoles, Ellie Pavlick, Quanze Chen, and Chris
  Callison{-}Burch. 2016.
\newblock Optimizing statistical machine translation for text simplification.
\newblock \emph{{TACL}}, 4:401--415.

\bibitem[{Zhang and Lapata(2017)}]{ZhangEA2017}
Zhang, Xingxing and Mirella Lapata. 2017.
\newblock Sentence simplification with deep reinforcement learning.
\newblock \emph{CoRR}, abs/1703.10931.

\bibitem[{Zhu, Bernhard, and Gurevych(2010)}]{ZhuEA2010}
Zhu, Zhemin, Delphine Bernhard, and Iryna Gurevych. 2010.
\newblock A monolingual tree-based translation model for sentence
  simplification.
\newblock In \emph{{COLING} 2010}, pages 1353--1361.

\end{thebibliography}
\end{CJK*}
\end{document}